\documentclass[11pt,a4paper]{article}
\usepackage{times,latexsym}
\usepackage{url}
\usepackage[T1]{fontenc}
\usepackage[acceptedWithA]{tacl2021v1}

\usepackage{tacl2021v1}

\usepackage{xspace,mfirstuc,tabulary}

\newif\iftaclinstructions
\taclinstructionsfalse % AUTHORS: do NOT set this to true
\iftaclinstructions
\renewcommand{\confidential}{}
\renewcommand{\anonsubtext}{(No author info supplied here, for consistency with
TACL-submission anonymization requirements)}
\newcommand{\instr}
\fi

\iftaclpubformat % this "if" is set by the choice of options

\else

\fi

\usepackage{algorithm}
\usepackage{algorithmic}
\usepackage{latexsym}
\usepackage{microtype}
\usepackage{graphics}
\usepackage{arydshln}
\usepackage{booktabs}
\usepackage{amsmath, amssymb}
\usepackage{soul}
\usepackage{diagbox}
\usepackage{graphicx}
\usepackage{float}
\usepackage{enumitem}
\usepackage{multirow}
\usepackage{todonotes}
\usepackage{lipsum}
\usepackage{bbding}
\usepackage{array}
\usepackage{soul}
\usepackage{appendix}
\usepackage{times}
\usepackage{latexsym}
\usepackage[T1]{fontenc}
\usepackage[utf8]{inputenc}
\usepackage{microtype}
\usepackage{inconsolata}
\usepackage{caption}
\usepackage{booktabs}
\usepackage{xcolor,colortbl}
\usepackage{newfloat}
\usepackage{listings}
\usepackage{amsmath,amsfonts}
\usepackage{soul}
\usepackage{pifont}
\usepackage{subcaption}  % For subfigures
\usepackage{pifont} 
\usepackage{tcolorbox}

\newcommand{\data}{\texttt{MENTAL-TRUST}}
\newcommand{\bench}{\texttt{TRUST-BENCH}}

\author{
  Aseem Srivastava$^1$, Zuhair Hasan Shaik$^2$, Tanmoy Chakraborty$^{3}$, Md Shad Akhtar$^1$
  \\
  $^1$IIIT Delhi, India \hspace{0.5cm} $^2$IIIT Dharwad, India \hspace{0.5cm} $^3$IIT Delhi, India\\
  \{\texttt{aseems}, \texttt{shad.akhtar}\}\texttt{@iiitd.ac.in}, \texttt{zuhashaik12@gmail.com}, \texttt{tanchak@iitd.ac.in}
}

\title{Trust Modeling in Counseling Conversations: A Benchmark Study}

\date{}

\begin{document}
\maketitle
\begin{abstract}
In mental health counseling, a variety of earlier studies have focused on dialogue modeling. However, most of these studies give limited to no emphasis on the quality of interaction between a patient and a therapist. The therapeutic bond between a patient and a therapist directly correlates with effective mental health counseling. It involves developing the patient's trust on the therapist over the course of counseling. To assess the therapeutic bond in counseling, we introduce {\em trust} as a therapist-assistive metric. Our definition of trust involves patients' willingness and openness to express themselves and, consequently, receive better care. We conceptualize it as a dynamic trajectory observable through textual interactions during the counseling. To facilitate trust modeling, we present \data, a novel counseling dataset comprising manual annotation of $212$ counseling sessions with first-of-its-kind seven expert-verified ordinal trust levels. We project our problem statement as an ordinal classification task for trust quantification and propose a new benchmark, \bench, comprising a suite of classical and state-of-the-art language models on \data. We evaluate the performance across a suite of metrics and lay out an exhaustive set of findings. Our study aims to unfold how trust evolves in therapeutic interactions.
\end{abstract}

\section{Introduction}
Mental health counseling relies on dynamic, evolving strategies tailored to the needs of each patient. For therapists, a successful session hinges on their ability to gauge the effectiveness of their approach, often relying on subtle cues to adjust their methods. This adaptability becomes especially crucial in text-based therapy, where the challenge of understanding patient engagement is even more complex. One of the many contributing factors responsible for effectiveness is the therapeutic bond. To assess this, we propose the need for a {\em therapist-assistive metric} to evaluate the success of ongoing strategies, one that could signal when recalibration is needed during counseling \cite{10.3389/fdgth.2022.847991}. In this work, we propose \textit{trust} as a metric to assess the therapeutic bond.
Trust, in a therapeutic context, reflects a patient’s willingness to disclose sensitive, personal matters; it unfolds as a dynamic trajectory rather than a static state. For instance, when a patient shifts from surface-level observations (e.g., “I’ve been feeling a little down”) to deeper self-exposure (e.g., “I’ve been scared of losing my job because I think it defines me”), it signifies growing trust. Conversely, hesitancy or disengagement might indicate faltering trust. The need for trust as an evaluative metric is particularly urgent in text-based therapy, which has seen exponential growth in recent years \cite{TrustnRespect}. However, they also place the burden of building and sustaining trust solely on the therapist’s linguistic and the patient's reactions \cite{li-etal-2023-understanding}. Current research in mental health focuses primarily on tracking sentiment, emotion, or engagement but lacks mechanisms to assess the interaction quality \cite{info:doi/10.2196/27868,info:doi/10.2196/46448,info:doi/10.2196/43271}.

Existing research has explored related areas like sentiments, user reactions, and moods in conversations \cite{info:doi/10.2196/30439, info:doi/10.2196/60589}. Still, it often falls short of addressing trust as a measurable construct \cite{info:doi/10.2196/jmir.8252}. Unlike existing attributes, trust possesses an evaluative component: it not only reflects the patient's feelings but also signals the effectiveness of the ongoing therapy approach. Without a system to identify trust flow, therapists and the evaluative systems designed to assist them are left without actionable insights into the relational dynamics of their sessions. 

In our work, we address this gap by proposing \textit{trust}, a novel and expert-verified metric designed to assess therapeutic bonds. We conceptualize trust as a patient’s willingness to disclose sensitive, relevant matters, reflecting their engagement and confidence in the therapeutic process. To practically assess this, we propose \data, a large, manually annotated dataset capturing trust dynamics and topic-switch patterns across patient-therapist dialogues. Furthermore, we evaluate the potential of computational models for trust detection through \bench, a comprehensive benchmarking of classical and state-of-the-art language models. Finally, we conduct an in-depth analysis of trust trajectories, uncovering key insights into how trust evolves during counseling conversations. 
Our findings reveal that while large language models (LLMs) excel at capturing topical understanding, they often struggle with accurately detecting fluctuations in trust levels. Interestingly, smaller models such as BART, BERT
and DeBERTa outperforms LLMs in this regard, showcasing a stronger ability to capture subtle variations in trust. Additionally, our in-depth analysis of trust trajectory paths, comparing increasing, decreasing, and neutral fluctuations, demonstrates that these smaller models effectively follow trust trajectories across all three cases. 
Our contributions are summarized below.

\begin{itemize}
 [leftmargin=*, noitemsep]
    \item We propose trust, a novel metric for evaluating therapeutic bonds.
    \item \data: We propose a novel and expert-backed annotated counseling dataset, capturing trust levels and topic-switch patterns in patient-therapist dialogues.
    \item \bench: We benchmark the performance of state-of-the-art language models, assessing their ability to detect trust.
    \item We analyze and criticize the findings from \bench\ to further understand trust trajectories, highlighting their potential to assist therapists in real-time strategy recalibration.
\end{itemize}

\paragraph{Reproducibility.} We commit to open-source the dataset for research purposes on final acceptance. 

\section{Related Work}
Mental health has emerged as a profound global concern, necessitating innovative solutions beyond traditional approaches. The rise of digital interventions, particularly
AI-powered tools, such as chatbots and diagnostic algorithms, have transformed the landscape of mental healthcare. Amidst growing demand and stretched resources, current research has focused on multiple verticals. Earlier works within dialogue space for counseling understanding include counseling understanding \cite{10.1145/3488560.3498509, srivastava2023criticalbehavioraltraitsfoster}, ethics \cite{Olawade2024EnhancingMH}, counseling note generation \cite{10.1145/3534678.3539187, srivastava2024knowledgeplanninglargelanguage}, behavior assessment, generation \cite{10.1145/3543507.3583380}. However, barely any work touches upon the quality of interaction between therapist and patient.

\paragraph{Study of bonds and psychology in digital interventions.}
The study of human bonds, especially in the context of psychological interactions \cite{merino2024body, park2024human, adhikary2024exploringefficacylargelanguage}, has gained increased attention in digital interventions. Research has shown that building a therapeutic bond between patients and healthcare providers is a critical factor in treatment outcomes. In traditional settings, this bond is forged through direct human interaction. However, as current methods become more prevalent in counseling, understanding how bonds form between patients and these systems has become essential. Studies highlight the challenges of replicating the empathetic connection that a human therapist provides, pointing out that AI tools often struggle to instill a sense of trust and rapport, both crucial elements of therapeutic progress \cite{kuhail2024human}.

\paragraph{Trust.} Plutchik's wheel of emotions is one the early successful attempts to classify human emotions into eight basic categories, including trust \cite{plutchik}. Here, trust is understood as the emotion that signals that `something is safe.’ Major theories in human psychosocial development emphasize the relevance of trust in building well-functioning relationships between individuals \cite{bowlby1969attachment, erikson1963childhood}, although there has been little research specifically on the patient’s level of trust toward psychotherapists \cite{birkhauer2017trust}. \citet{carter2024therapeutic} attempted to define therapeutic trust (default therapeutic trust and overriding therapeutic trust), and differentiated it from ordinary trust. \citet{crits2019trust} aimed to develop and report preliminary psychometric analyses of a new brief measure to evaluate a patient’s level of trust and respect for their clinician. However, none of these model the trust dynamics of the patient during the progress of a dialogue.

In contrast to previous definitions that conceptualize trust as a static measure, our work focuses on trust as a dynamic trajectory within dialogue. Unlike psychometric measures that evaluate trust as an outcome, we define trust as a moment-by-moment phenomenon, capturing its fluctuations in response to the patient-therapist interaction. 

\begin{table*}
\centering
\resizebox{\textwidth}{!}{%
\begin{tabular}{lp{45em}cp{7em}} \toprule
\#&\textbf{Utterances} & \textbf{Trust} & \textbf{Topic Discussing} \\
\midrule
1& \underline{Therapist:} In the other chair, let's have critic that's inside, that internal critic. That's advocating for settling down, for getting married, for having kids. \colorbox{green!26}{In a way, it represents societal pressure}. These two sides separate to you? \newline
\underline{Patient:} Yeah. & 1.5 & Balancing career and settling down\\ 
\midrule
2& \underline{Therapist:} You feel like \colorbox{green!26}{you can be your emotional self here and the internal critic there?} \newline
\underline{Patient:} \colorbox{green!26}{I think so.} & 1.5 & Balancing career and settling down\\
\midrule
3& \underline{Therapist:} Ok so, I think you remember how this works. You'll be talking to the internal critic, right? So, the internal critic is on the side of settling down and getting married. And then, when you're ready to have the critic respond, you'll switch to that chair and respond to the emotional, needs-driven side. Does that make sense? \colorbox{green!26}{So whenever you're ready, go ahead and speak directly to the internal critic.}\newline
\underline{Patient:} I want to finish what I'm doing, and I want to finish what I started. I've worked so hard on my career path, and I'm so close to finishing. I want to finish for me. \colorbox{green!26}{I want to accomplish those goals that I've set for myself.} It's something I'm passionate about. & 2 & Balancing career and settling down\\
\midrule
4& \underline{Therapist:} So you've made your statement. If you are ready to switch then I use the internal critic. \newline
\underline{Patient:} There's always time to pursue a career, but I need to focus on what's expected of me at this time, on the best age for having kids I have a loving boyfriend who is ready and wants to get married and settle down. \colorbox{green!26}{And you are being selfish by just following a path that you can take at any time.}& 2 & Balancing career and settling down\\

\midrule
5&\underline{Therapist:} Okay. So the internal critic is just called a selfish and one of the things is a response. \newline
\underline{Patient:} I feel like I'm being attacked. \colorbox{green!26}{And it's not wrong to think about myself} and what I am passionate about. & 2.5 & Balancing career and settling down\\
\midrule
6& \underline{Therapist:} Let me interrupt you. So your hands are pointing back toward the career emotional self. What can you do with \colorbox{green!26}{your hands that really embody that you want to communicate to the internal critic?} Anything else?  \newline
\underline{Patient:} The pointing towards me to protect myself from what critic is saying. I wanna push that back to the critic.
& 3 & Decision Making\\

\bottomrule

\end{tabular}
}
\caption{Dataset example. The table showcases a dialogue snippet from the \data\ dataset, accompanied by trust level annotations and the discussion topic. We highlight the key portions of the text contributing significantly to the constructive and incremental trust trajectory for better interpretability.}
\label{tab:exampleDialogMainPaper}
\vspace{-3mm}
\end{table*}

\section{Dataset}
We propose \data, a trust-level rated counseling dataset specifically curated for trust modeling in counseling conversations. The dataset consists of $12.9K$ utterances extracted from 212 dialogues, where each patient utterance is annotated with seven expert-verified ordinal trust labels. These labels are designed to capture the nuanced progression of trust within therapeutic interactions. To create \data, we extend the publicly available HOPE counseling dataset, which includes a total of 12,912 utterances across multiple counseling sessions \cite{10.1145/3488560.3498509}. Our subset focuses exclusively on patient utterances, providing a targeted framework for studying trust as a dynamic construct. The following subsections detail the data collection process, annotation scheme, trust labels, and key dataset statistics.

\subsection{Data Annotation}
Counseling conversations differ from standard dialogues, particularly in their nuanced focus on emotional depth, trust dynamics, and therapeutic context \cite{10038072,10.1145/3581783.3612346,chen2024convcounselconversationaldatasetstudent}. These unique characteristics necessitate a specialized annotation approach to capture the intricate patterns. To address this, we collaborate with mental health experts to design a specialized annotation scheme. This scheme comprises seven ordinal trust levels, explicitly crafted to capture the evolving dynamics. Before detailing the annotation guidelines, it is essential to first establish an understanding of trust, which we discuss next.

\paragraph{Trust.} Trust reflects the patient's willingness to depend, or intent to depend, on the expert/therapist with a feeling of relative security in spite of a lack of control over the expert, even though negative consequences are possible. Specifically, the trust of a patient towards their therapist is characterized by (a) sharing personal, detailed, or sensitive information, (b) opening up about relevant concerns, and (c) maintaining focus on the topic of concern without unnecessary deviation. 
It is worth noting that our primary goal is to assess the patient's trust in the therapist; however, the impact of the therapist’s responses and interventions would affect the client’s degree of opening up.
Therefore, we annotate trust levels for every patient utterance from seven levels defined by experts.
A trust level is an ordinal value assigned to each flip, determined by considering its conversational context and preceding trust levels. Finally, prior to annotation, the relevant topic of discussion is identified to ensure accurate context alignment. To make it more streamlined and precise to ongoing annotations, we mark the change of topic, both instances and the shifted topic. Annotators evaluate and update trust levels based on shifts in the conversation’s focus for better alignment with the identified topic. Detailed annotation guidelines are discussed below.

\subsection{Annotation Guidelines}
To effectively capture the dynamics of trust in counseling conversations, we develop a framework of seven ordinal trust levels. These levels represent a spectrum of a patient's openness, ranging from complete refusal to engage with the therapist to fully opening up about their core issues. While there are seven levels in total, four major levels: least trust (L1), low trust (L2), building trust (L3), and achieved trust (L4), serve as the foundation. The intermediate levels (e.g., 1.5, 2.5, and 3.5) account for complex situations where a patient’s trust behavior does not fit neatly into one of the major categories. Below, we define each trust level:

\begin{enumerate}[leftmargin=*, noitemsep]
    \item {\bf Least Trust.} This level is assigned when the client demonstrates an active refusal to open up. This could include non-aligning responses or outright rejection of the therapist's attempts to engage.
    \item {\bf Low Trust.} This level represents a slight increase in trust from Level 1 or a slight decrease from Level 3. The patient shows hesitation when opening up but does not completely refuse. Indicators of low trust include:
    \begin{itemize}[leftmargin=*, noitemsep]
        \item Limited self-expression.
        \item Use of discussion filler such as “um” or “hm”.
        \item Expressions of insecurity or doubtfulness, such as “maybe” or “I guess”.
    \end{itemize}

    \item {\bf Building Trust.} This level is scored when the patient demonstrates consistent engagement and openness, though they may still require prompting from the therapist or digress from the main topic of concern.

    \item {\bf Achieved Trust.} This level is assigned when the patient fully opens up to discuss their core issues without any digression. At this point, they actively engage with the therapist on the point of concern, indicating complete trust.
\end{enumerate}

\paragraph{Intermediate Trust Levels.} In many conversations, patient utterances do not always fit neatly into one of the four primary trust levels. For instance, an utterance might exhibit elements of both Level 1 (least trust) and Level 2 (low trust), creating ambiguity for the annotators. To address this challenge, we experiment with intermediate levels, such as 1.5, 2.5, and 3.5, positioned between the major trust levels. These midpoints allow annotators to capture complex shifts in trust that might otherwise lead to inconsistencies. Not only annotators report that these additional levels significantly improved their ability to assign accurate trust scores, resulting in better alignment with expert judgment and inter-annotator agreement. One such importance is that the conversations are initialized with a trust level of $2.5$, representing a neutral midpoint. This is a fair assumption to ensure consistency across annotations, as there is no prior context available to set a more specific initial trust level. From this point, annotators observe the patient’s trust trajectory and annotate based on the guidelines.

\paragraph{Annotating Topic Shifts.}
In addition to trust levels, we annotate topic shifts to capture changes in the focus of the conversation. A topic shift is marked whenever the conversation deviates from the current topic of discussion to introduce a new subject. This annotation helps track how trust dynamics interact with the flow of dialogue, as maintaining alignment. We ask annotators to record the new topic introduced and the point in the dialogue where the shift occurred, ensuring a detailed record of conversational transitions and {\em digressions}.

We present a dialogue snippet in Table \ref{tab:exampleDialogMainPaper} to showcase example annotations. Along with the dialogue, it includes the topic and trust level annotations, highlighting key factors contributing to incremental trust levels. We include additional examples in Appendix (c.f. Table \ref{tab:exampleDialog}).

\begin{figure}
    \centering
    \includegraphics[width=\columnwidth]{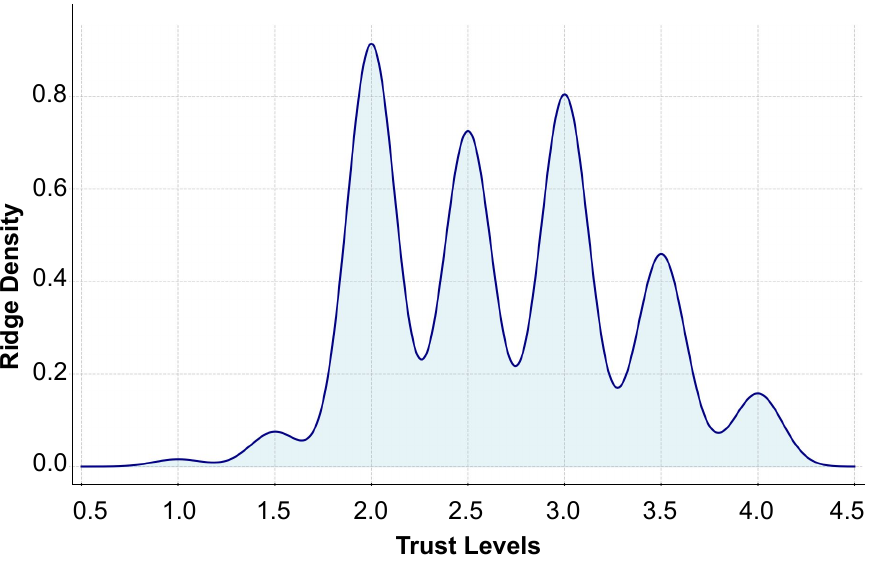}
    \caption{Distribution of ridge density for each trust level between 1 (min) -- 4 (max) in \data.}
    \label{fig:ridgeDen}
    \vspace{-4mm}
\end{figure}

\begin{table*}[ht]
\small
\centering
\resizebox{\textwidth}{!}{
\begin{tabular}{lccccccc}
\toprule

\multirow{2}{*}{\bf Models} & \multirow{2}{*}{\bf \#Sessions}  & \multirow{2}{*}{\bf \#Utterances} & \multirow{2}{*}{\bf Avg. Utterance Length} & \multicolumn{2}{c}{\bf Avg. Utt. Len./Spk.} & \multicolumn{2}{c}{\bf Avg. Tok/Spk./Utt.} \\
\cmidrule{5-8}
&&&&\bf T &\bf P &\bf T &\bf P \\
\midrule
Train   & 116 & 6902 & 56.56 & 28.51 & 28.05 & 26.15 & 29.75 \\
Test    & 17 & 949 & 52.71 & 27.00 & 25.71 & 25.07 & 21.06 \\
Val     & 34 & 2321 & 65.21 & 32.71 & 32.50 & 26.73 & 19.32 \\
\bottomrule
\end{tabular}
}
\caption{Split-wise dialogue- and speaker-based analysis of the proposed dataset, \data.}
\label{table:stats}
\end{table*}

\subsection{Annotation Process}
To ensure reliable annotations, we adopt an iterative approach involving three expert linguists and a team of domain experts. In Iteration 1, we begin with four primary trust levels only for a random sample of 10 dialogues and ask them to assign trust levels based on the initial guidelines. However, this iteration revealed significant disagreements, with an inter-annotator agreement (IAA) score of $0.22$, as many utterances exhibited mixed characteristics that did not fit neatly into the defined levels. In Iteration 2, we introduced intermediate trust levels, and this refinement initially resolved much of the disagreement. However, a new challenge emerged: subjective interpretations of whether a patient’s utterance reflected engagement with the relevant topic of concern or constituted digression. Despite this, the IAA score improved to $0.43$. To address this issue, Iteration 3 incorporated expert-provided criteria to clarify how the topic of discussion should influence trust levels. Annotators were instructed to consider whether utterances aligned with the core topic or diverged as digressions. With these refined guidelines, annotations became more consistent, and the final iteration achieved a Cohen’s Kappa score of $0.77$, falling into the `substantial’ agreement category. This iterative process not only resolved key ambiguities but also ensured a robust and reliable annotation framework for the dataset.

\section{Data Analysis}
We conduct a detailed analysis examining both the structural composition and trust dynamics to understand the characteristics of \data.

\paragraph{Statistical Analysis.}
Table \ref{table:stats} presents the key statistics of our dataset split across training, testing, and validation sets. The dataset comprises a total of $167$ counseling sessions, with $116$ sessions ($6,902$ utterances) allocated to training, $17$ sessions ($949$ utterances) to testing, and $34$ sessions ($2,321$ utterances) to validation. The average utterance length shows consistent patterns across splits, with validation utterances being slightly longer ($65.21$ tokens) compared to training ($56.56$ tokens) and test ($52.71$ tokens) sets. This variation is natural in therapeutic conversations where utterance length often correlates with the depth of discussion. 

We observe interesting patterns in speaker-specific assessment. The average utterance length per speaker shows therapists maintaining relatively consistent lengths across splits ($28.51, 27.00,$ and $32.71$ tokens), while patients demonstrate similar consistency ($28.05, 25.71,$ and $32.50$ tokens). This balance between therapist and patient utterance lengths suggests equitable dialogue participation. Likewise, the average tokens per speaker per utterance reveals that therapists typically use fewer tokens ($26.15, 25.07,$ and $26.73$) compared to patients in the training and test sets ($29.75$ and $21.06$), except in the validation set where patient utterances are notably more concise ($19.32$). This pattern might reflect varying degrees of patient expressiveness across different sessions, while therapists maintain more consistent communication patterns.

\begin{table*}[t]
\centering
\resizebox{\textwidth}{!}{

\begin{tabular}{lcccccccccc}
\toprule
% \multirow{2}{*}{Models} & \multicolumn{3}{c}{Disorder Classification} & \multicolumn{4}{c}{Classification Metrics} & \multirow{2}{*}{Ordinal-Crossentropy} & \multirow{2}{*}{MSE} \\
% \cmidrule(lr){2-4} \cmidrule(lr){5-8}
\textbf{Models} & Acc & $F1_m$ & $F1_w$ & $P_m$ & $R_m$ & $P_w$ & $R_w$ & $OL_{CE}$  & Params \\
\midrule
\textbf{{Encoder Only}} \\
\quad BERT \cite{BERT} & 85.04 & 68.93 & 84.90 & 69.02 & 69.16 & 85.02 & 85.04 & 3.67 & 109 M \\
\quad Mental-BERT \cite{mentalBERT}& 84.12 & 70.14 & 83.87 & 71.83 & 69.12 & 83.99 & 84.12 & 3.75 & 109 M \\
\quad RoBERTa \cite{liu2019robertarobustlyoptimizedbert}& 80.76 & 61.96 & 81.01 & 61.58 & 63.60 & 82.12 & 80.76 & 4.28 & 124 M \\
\quad ALBERT \cite{lan2020albertlitebertselfsupervised}& 84.48 & 72.77 & 84.34 & 83.88 & 70.21 & 85.19 & 84.48 & 3.73 & 11 M \\
\quad DeBERTa \cite{he2021debertadecodingenhancedbertdisentangled} & 85.30 & 72.01 & 85.23 & 82.82 & 72.91 & 85.81 & 85.30 & 3.46 & 184 M \\
\quad XL-Net \cite{yang2020xlnetgeneralizedautoregressivepretraining}& 78.76 & 53.49 & 78.22 & 52.60 & 55.26 & 78.27 & 78.76 & 4.89 & 117 M \\
\midrule

\textbf{{Encoder-Decoder}} \\
\quad BART \cite{lewis2019bartdenoisingsequencetosequencepretraining}& 87.39 & 81.15 & 87.34 & 84.82 & \textbf{79.35} & 87.52 & 87.39 & 2.92 & 407 M \\
\quad Mental-BART \cite{yang2023mentalllama} & \textbf{89.03} & \textbf{81.63} & \textbf{88.96} & \textbf{89.48} & 78.22 & \textbf{89.15} & \textbf{89.03} & \textbf{2.53} & 407 M \\
\quad T5 \cite{2020t5} & 71.82 & 49.58 & 70.93 & 51.04 & 49.70 & 71.10 & 71.82 & 5.93 &  738 M \\
\midrule

\textbf{{Decoder Only}} \\
\quad Llama 3.1  \cite{llama3} & 66.51 & 43.72 & 64.63 & 48.50 & 42.83 & 65.92 & 66.51 & 6.60 & 8000 M \\
\quad Mistral-7b \cite{mistral} & 64.17 & 42.60 & 62.75 & 47.36 & 42.13 & 64.13 & 64.17 & 7.11 & 7000 M \\
\quad Phi 3.5 \cite{phi3} & 74.43 & 50.24 & 73.10 & 52.98 & 49.52 & 73.27 & 74.43 & 5.27 & 3800 M \\
\midrule
\textbf{{Closed-Source}} \\
\quad GPT 4o \cite{petruzzellis2024benchmarkinggpt4algorithmicproblems}& 23.38 & 12.85 & 20.57 & 14.80 & 15.38 & 24.37 & 23.38 & - & Large \\
\quad Gemini 1.5 \cite{geminiteam2024gemini15unlockingmultimodal}& 23.58 & 12.06 & 19.10 & 15.79 & 17.56 & 25.38 & 23.58 & - & Large \\
\bottomrule
\end{tabular}
}
\caption{{TrustBench -- Benchmarking.} We evaluate 14 models on the proposed dataset, \data, across four major categories: encoder-only, decoder-only, encoder-decoder, and closed-source. Evidently, smaller models outperform larger models across eight metrics, including Accuracy (Acc), F1, Precision (P), Recall (R), and Ordinal Cross-Entropy Loss ($OL_{CE}$).}

\label{tab:model_comparison}
\end{table*}

\paragraph{Trust Distribution.} Analysis of trust level distribution in our dataset reveals notable patterns in therapeutic trust dynamics. Figure \ref{fig:ridgeDen} presents a ridge density plot of trust levels, demonstrating three distinct peaks. The primary peak occurs at trust level $2.0$ with the highest density of $0.88$, indicating that moderate trust levels are most prevalent. Two secondary peaks emerge at trust levels $2.5$ and $3.0$, suggesting that patients commonly transition through these intermediate trust states. On the other hand, the density tapers significantly at both extremes (below $1.0$ and above $4.0$), indicating that both very low and very high trust states are relatively rare. This aligns with clinical observations that establishing deep trust requires sustained therapeutic engagement, while complete distrust is uncommon once patients initiate therapy.

\section{TrustBench}
Building on our definition of trust, we propose TrustBench, a benchmarking framework designed to evaluate how well existing methods can quantify trust in counseling conversations. We operationalize the problem statement as an ordinal classification task. This section outlines the methodologies, experiments, and evaluation metrics employed in TrustBench to benchmark state-of-the-art models.

\subsection{Trust Modeling}
The input to these models is meant to capture the dynamics of trust through conversational context rather than treating utterances in isolation. To capture the dynamics of trust as it evolves during a conversation, we consider contextual representations of patient utterances within their dialogue history. Given a dialogue $D$ consisting of $n$ utterances $\{u_1, u_2, \dots, u_n\}$, where each utterance $u_i$ is either from the patient ($P$) or the therapist ($T$), the task is to predict the trust score $t_i \in \{1, 2, 3, 4, 5\}$ for each patient utterance $u_i$. The context window $\{u_{i-k}, \dots, u_{i-1}\}$, consisting of the $k$ preceding utterances for context.

\subsection{Competing Models}
We discuss 14 competitive yet diverse baseline methods that we employ to benchmark. 
\paragraph{Enoder-only Methods.} $\blacktriangleright$\textbf{BERT} \cite{BERT} is a bidirectional model that excels in understanding context through deep pretraining on large text corpora. $\blacktriangleright$\textbf{Mental-BERT} \cite{mentalBERT} is a variant fine-tuned on mental health data, making it well-suited for tasks with psycholinguistic nuances. $\blacktriangleright$\textbf{RoBERTa} \cite{liu2019robertarobustlyoptimizedbert} improves upon BERT by optimizing training strategies, making it highly effective for a wide range of NLP tasks. $\blacktriangleright$\textbf{ALBERT-base-v2} \cite{lan2020albertlitebertselfsupervised} is a lighter version of BERT that reduces model complexity while maintaining competitive performance. $\blacktriangleright$\textbf{DeBERTa-v3-base} \cite{he2021debertadecodingenhancedbertdisentangled} incorporates disentangled attention mechanisms, allowing it to better capture semantic relationships. $\blacktriangleright$\textbf{XLNet} \cite{yang2020xlnetgeneralizedautoregressivepretraining} uses a permutation-based training approach, which enhances its capability to model bidirectional context.

\begin{figure*}[t]
  \centering
  % \scalebox{1.01}{
  \includegraphics[width=\textwidth]{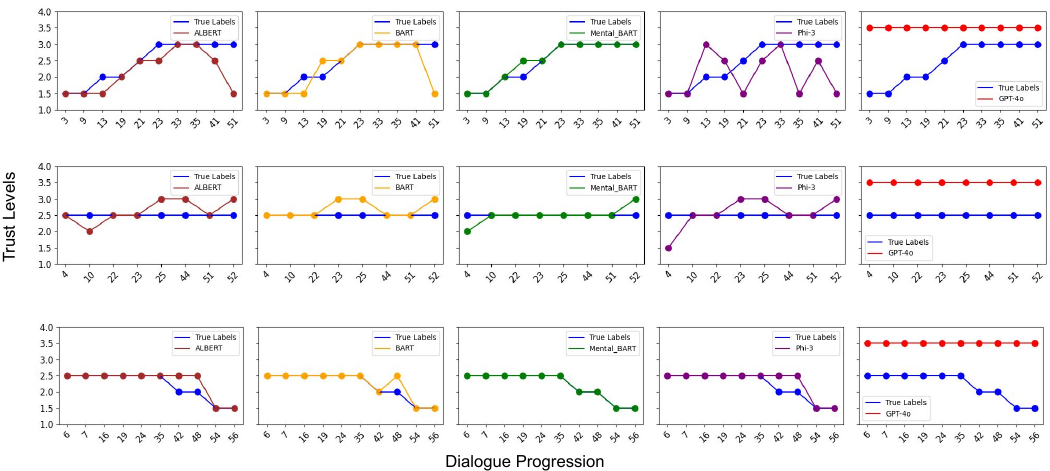}
  % }
    \caption{Examples of trust progression scenarios (\textit{increasing}, \textit{constant}, \textit{decreasing}) across top-performing models. The ground-truth trust values are represented by the blue line. Decoder-only models, particularly domain-specific variants such as \text{Mental-BART}, achieve the closest alignment with true trust values, while closed-source models perform the worst due to their inherent rigidity.}
  \label{fig:trustTrajectory}
\end{figure*}

\paragraph{Encoder-Decoder Methods.}$\blacktriangleright$\textbf{BART} \cite{lewis2019bartdenoisingsequencetosequencepretraining} is pre-trained as a denoising autoencoder. $\blacktriangleright$\textbf{Mental-BART} \cite{yang2023mentalllama}, fine-tuned on mental health conversations, makes it highly effective for generating therapeutic dialogue. $\blacktriangleright$\textbf{T5} \cite{2020t5} reframes NLP tasks into a text-to-text format, making it a versatile encoder-decoder model capable of diverse tasks.

\paragraph{Decoder-only Models.} $\blacktriangleright$\textbf{LLaMA-3.1} \cite{llama3} is designed for efficiency with smaller datasets, demonstrating strong performance. $\blacktriangleright$\textbf{Mistral-7B-v0.1} \cite{mistral} is optimized for language modeling. $\blacktriangleright$\textbf{Phi-3.5} \cite{phi3} is a small language model (SLM) particularly useful for easy compute power and inference while maintaining the abilities of LLMs. 

\paragraph{Close-sourced Models.} We also integrate closed-source, state-of-the-art models such as $\blacktriangleright$\textbf{GPT-4o} \cite{petruzzellis2024benchmarkinggpt4algorithmicproblems}, known for its complex reasoning and strong performance in general NLP and instruction following tasks, and $\blacktriangleright$\textbf{Gemini} \cite{geminiteam2024gemini15unlockingmultimodal}, which excels in multitasking scenarios, particularly in generative AI.

\section{Result and Analysis}
Here, we discuss the performances achieved by participating methods.

\subsection{Performance Comparison}
We benchmark the proposed dataset, \data, evaluating the performance of 14 competing baseline methods across nine evaluation metrics. The baseline methods are categorized into four major architectures: encoder-only, encoder-decoder, decoder-only, and closed-source language models. Table \ref{tab:model_comparison} highlights Mental-BART as the best performer.

Within the encoder-only category, DeBERTa achieves the best results, surpassing other methods with scores of 85.30 (accuracy), 85.23 ($F1_w$), 85.81 ($P_w$), and 3.46 ($OL_{CE}$). Notably, DeBERTa’s performance closely aligns with that of ALBERT, demonstrating competitive capabilities in this architecture. On the other hand, for the decoder-only methods, Phi-3.5, a small language model, emerges as the best performer, recording scores of 74.43 (accuracy), 73.10 ($F1_w$), 73.27 ($P_w$), and 5.27 ($OL_{CE}$). Meanwhile, in zero-shot comparisons with closed-source LLMs such as GPT-4o and Gemini-1.5, these models fall short of fine-tuned methods across all metrics. 
Surprisingly, despite their advanced capabilities, even these sophisticated models fail to outperform smaller, fine-tuned encoder-only models like DeBERTa.

\begin{figure}[!t]
  \centering
  \scalebox{0.8}{\includegraphics[width=\columnwidth]{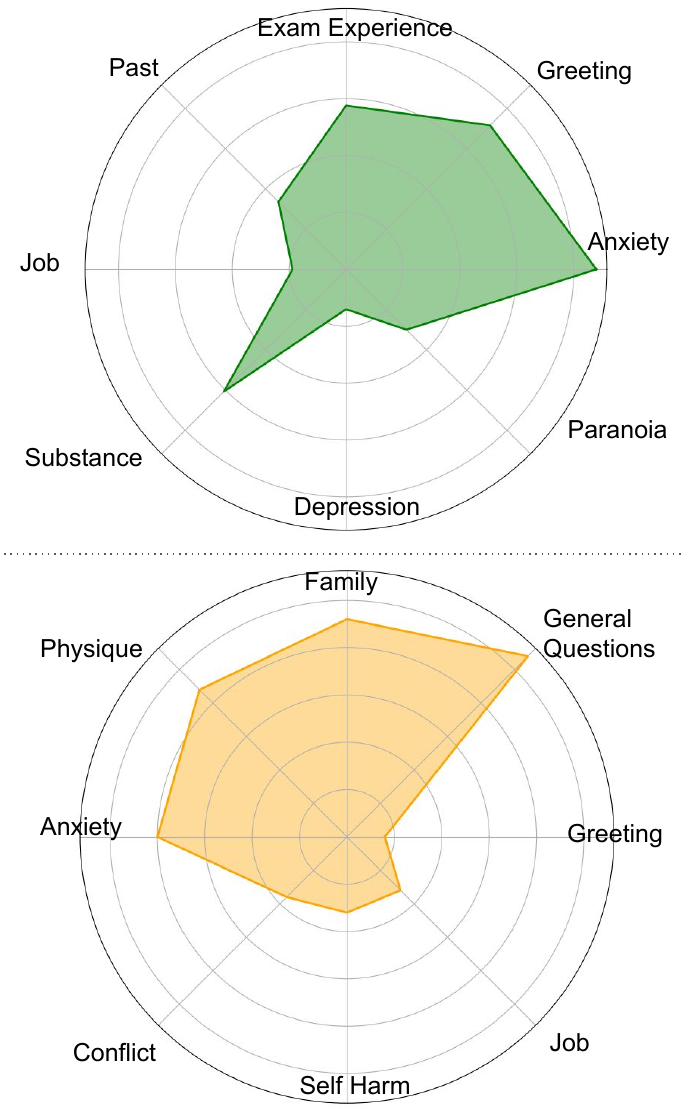}}
  \caption{Topic of discussion in counseling sessions segmented into two major segments. Top: illustrating core topics in sessions with a positive outcome; Bottom: showing topics in sessions with negative outcomes.}
  \label{fig:page1example}
\end{figure}

Finally, we test full-fledged transformer architectures, including BART and its variant Mental-BART. These models consistently outperform all other baselines across all the segments, with Mental-BART achieving the best scores on seven out of nine metrics. For Mental-BART, we observe a performance of 89.03 (accuracy), 88.96 ($F1_w$), 89.15 ($P_w$), and 2.53 ($OL_{CE}$), further cementing its position as the top-performing model on the proposed dataset.
Our findings underline the effectiveness of specialized and smaller models like Mental-BART in the trust modeling task and highlight the limitations of general-purpose LLMs in addressing the ordinal classification tasks. We discuss the effect of model sizes for trust modeling in Section \ref{smallVSlarge}.

\subsection{Analysis}
We study the dynamics of trust modeling in counseling, examining two major fronts: (a) trust trajectory modeling and (b) topical analysis.  We study these aspects for two therapeutic outcomes: {\em positive outcome}, where concluding trust exceeds initial trust levels and {\em negative outcome}, where concluding trust falls below initial trust levels. We layout evidential analysis and present our findings, highlighting both the strengths and limitations of models in capturing trust dynamics.

\subsubsection{Trust Trajectory Analysis}
The quantitative evaluation of trust modeling, as presented in Table \ref{tab:model_comparison}, provides a comprehensive overview of each model’s performance when compared with expert-annotated trust levels as an ordinal classification task. While these scores reflect the overall ability of models to predict trust levels, we further study to understand how well models follow trust trajectories from the beginning of a session. To achieve this, we analyze model-predicted trajectories against gold-standard annotations for three distinct cases based on their overall therapeutic outcome, including positive, negative, and neutral outcomes (where trust levels remain relatively stable).

To balance thoroughness and efficiency, we focus on the best-performing model from each architecture paradigm, including LLMs, SLMs, encoder-only, decoder-only, and transformer-based. Evidently, as shown in Figure \ref{fig:trustTrajectory}, Mental-BART demonstrates superior alignment with the gold standard, particularly excelling in scenarios involving abrupt fluctuations. Its performance closely rivals ALBERT, reflecting robustness in handling moderate trust variations. On the other hand, ALBERT, while competitive, shows limitations in scenarios involving sharp trust fluctuations, such as rapid dips or surges. Similarly, the SLM, Phi-3.5, struggles with rapid directional changes. 

\subsubsection{Topic vs Therapeutic Outcome}
We perform a topical analysis of counseling sessions to investigate thematic trends across different therapeutic outcomes. Figure \ref{fig:page1example} illustrates the frequency distribution of key topics discussed during these sessions, providing insights into the relationship between topic alignment and trust trajectories.

\begin{table*}[t]
\small
\centering
\resizebox{\textwidth}{!}{
\begin{tabular}{lccccccccc}
\toprule
\textbf{Splits} & \textbf{$+$ve Jumps} & \textbf{Max$(+\Delta)$} & \textbf{Min$(+\Delta)$} & \textbf{Avg($+\Delta)$} & \textbf{$-$ve Jumps} & \textbf{Min$(-\Delta)$} & \textbf{Max$(-\Delta)$} & \textbf{Avg$(-\Delta)$} & \textbf{Avg Streak} \\
\midrule
Train & 228 & 1.0 & 0.5 & 0.504 & 110 & -0.5 & -2.0 & -1.032 & 8.79 \\
Test  & 37  & 0.5 & 0.5 & 0.5   & 16  & -0.5 & -2.0 & -1.063 & 7.61 \\
Val   & 77  & 1.0 & 0.5 & 0.506 & 36  & -0.5 & -2.0 & -1.069 & 9.30 \\
\bottomrule
\end{tabular}
}
\caption{{Trust jumps and streak analysis.} The table shows trust dynamics and includes the counts of trust level jump, their respective maximum, minimum, and average changes ($+\Delta\ \text{and} -\Delta$), and the average streak length, highlighting the progression and fluctuations in trust trajectories.}
\label{tab:jumpAnalysis}
\end{table*}

Sessions with positive outcomes are characterized by `in-depth discussions' on focused topics such as {\em anxiety}, {\em academic examinations}, and {\em substance} use or abuse. These conversations display `minimal digressions', with patients maintaining alignment with the core topic of concern. Additionally, a recurring pattern emerges where sessions often begin or conclude with proper {\em greetings}, a seemingly routine element that contrasts sharply with negative outcome sessions, which frequently lack such exchanges. Furthermore, the topics discussed in positive outcomes tend to be coherent and closely aligned with the therapeutic goal. Conversely, sessions with negative outcomes exhibit scattered and tangential discussions, often touching on topics such as {\em general questions}, {\em casual chitchat}, {\em family} dynamics, {\em physical appearance}, and {\em anxiety}. A striking feature of these sessions is the prevalence of topic digressions, where patients shift focus away from the core concern to peripheral or unrelated issues, and an even bigger concern is `frequent digression'. These deviations are commonly linked to diminished trust, as the therapeutic bond is weakened when alignment on core emotional or therapeutic goals is not maintained.

\section{Discussion}
In this section, we discuss two key aspects of our work: (a) the patterns of trust evolution and  (b) the counterintuitive relationship between model size and trust modeling.

\subsection{Discussion on Effect of Model Sizes} \label{smallVSlarge}
Our findings reveal a counterintuitive relationship between model size and trust modeling performance. Despite the common assumption that LLMs lead to better performance, our analysis shows an inverse trend in the context of therapeutic trust modeling. Smaller encoder models like BERT ($109M$ parameters) and ALBERT ($11M$ parameters) achieve notably higher accuracy ($85.04\%$ and $84.48\%$ respectively) compared to larger models such as Llama 3.1 ($8B$ parameters) and Mistral ($7B$ parameters), which achieve only $66.51\%$ and $64.17\%$ accuracy respectively. This inverse relationship becomes even more surprising with closed-source models like GPT-4 and Gemini 1.5, which, despite their large parameter counts, perform significantly worse with accuracies around $23\%$. 

Our understanding of this pattern is that smaller models are more focused architectures suited to learning the specific patterns and subtle indicators of trust levels. For instance, smaller models are better at learning local, contextual patterns specific to trust dynamics rather than relying on broad world knowledge. Moreover, the specialized nature of trust modeling might benefit from models that can be more precisely fine-tuned to the task.

\begin{figure*}[t]
  \centering
  \includegraphics[width=\textwidth]{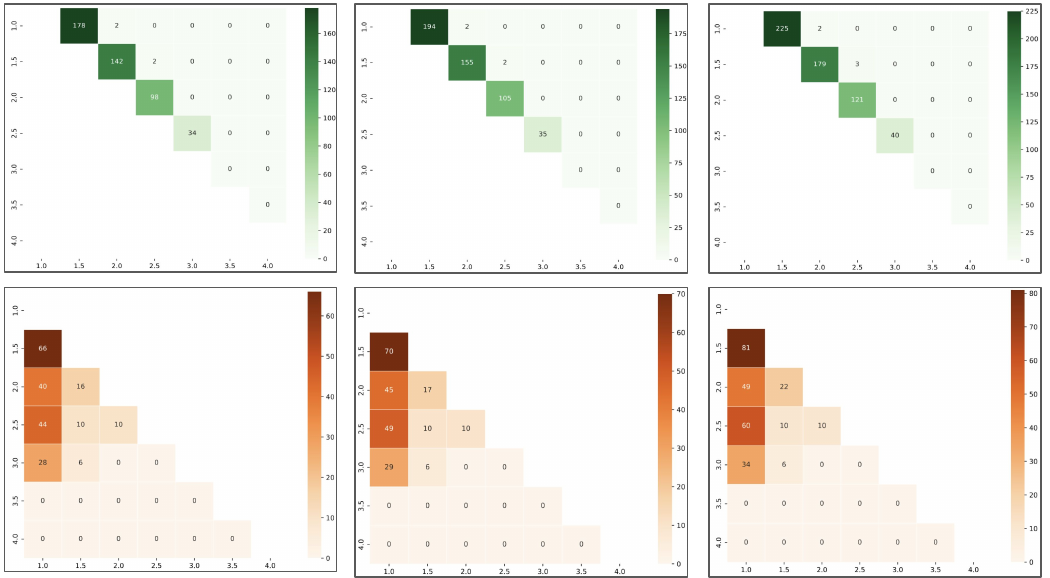}
  \caption{Illustration of the progression in trust scores, represented as either increasing or decreasing jumps. The y-axis denotes the starting trust score ("From"), while the x-axis indicates the resulting trust score ("To"). For instance, a progression from 2.0 to 2.5 corresponds to the (4, 5) cell in the upper triangular tables. Increasing trust score jumps are shown in green, while decreasing jumps are displayed in orange. Darker shades signify higher counts of the respective trust score changes.}
  \label{fig:confusionMat}
\end{figure*}

\subsection{Discussion on Trust Evolution Patterns}
The trust level typically progresses through three stages: {\em refusal} (L1-L2), {\em digression} (L2-L3), and {\em opening-up} (L3-L4). Refusal involves reluctance or rejection of the therapist’s prompts. Digression occurs when peripheral topics are discussed instead of the core issue, and opening-up reflects direct engagement with the concern. For instance, a woman from \data\ initially dismisses discussing resentment toward her mother-in-law (refusal), later shares unrelated grievances (digression), and eventually addresses her hurt over personal comments (opening-up). This behavior is unique to this problem -- trust modeling, and we further analyze this through the statistical point of view, as shown in Table \ref{tab:jumpAnalysis}. 

Our analysis yields distinct patterns in how trust evolves during therapeutic conversations. Trust progression demonstrates an inherent positive inclination, with positive transitions occurring approximately $2\times$ as negative ones, as building trust is a fundamental goal. Notably, \textit{we observe that trust typically builds gradually}, with average positive changes of \string~$(+0.5)$ levels, while trust decreases tend to be more dramatic, with average \string~$(-1.0)$ levels. This asymmetry suggests that while trust is carefully cultivated in therapeutic relationships, it can be easily diminished by situational digression.

Another interesting finding is the presence of trust stability periods, averaging $8-9$ consecutive utterances without level changes (c.f. {\em avg. Streak} in Table \ref{tab:jumpAnalysis}). These {\em trust plateaus} may represent stable therapeutic states where the patient-therapist bond has reached a temporary equilibrium. Figure \ref{fig:confusionMat} shows such plateaus in detail.

\section{Limitations and Future Work}
While our work studies trust modeling in counseling conversations, several limitations highlight opportunities for future research. The dataset is limited to interactions in English-language only, restricting its applicability in multilingual contexts and overlooking cultural nuances critical to therapeutic dynamics. We plan to address such gaps in future works by expanding the dataset to include diverse languages and cultural contexts. Additionally, the reliance on textual data alone omits other modalities, such as vocal tone and facial expressions, which are vital for capturing the full spectrum of trust dynamics. Incorporating these modalities into future models is vital for real-world applicability. Furthermore, we benchmarked various models but did not propose a method for trust modeling, which we plan to focus on in the future.

\section{Ethical Considerations}
Our work focuses on trust modeling with the primary objective of assisting therapists in developing more effective and adaptive therapeutic strategies. Importantly, this study is not patient-facing, ensuring that our study remains risk-free for patients. The data we use in this study is derived from a publicly available counseling dataset, which has already undergone necessary procedures to ensure user anonymity and protection of personally identifiable information (PII). Furthermore, every step of data construction and annotation has been carried out in close consultation with domain experts to ensure reliability and alignment with ethical standards. Despite the ethical practices we followed, we acknowledge the possibility of cultural biases due to its limited size and diversity. We explicitly address this limitation in our work and view it as an area for improvement in future iterations. Our study adheres to established ethical standards in data handling, privacy protection, and annotation.

\section{Conclusion}
In this paper, we introduced the novel concept of using trust as a metric to evaluate therapeutic strategies in text-based counseling. Our approach to quantifying trust dynamics leverages psycholinguistics and psychological theories, providing an actionable tool for therapists to calibrate their methods in real-time. We proposed a robust, annotated dataset containing trust ratings and topic-switch patterns from real counseling dialogues, which we used to benchmark state-of-the-art AI models. Our experiments demonstrated the viability of using AI to detect and quantify trust in therapeutic contexts, showing promising results in predicting patient engagement. The insights gained from this work lay the foundation for future advancements in automated mental health counseling support.

\bibliography{refs}
\bibliographystyle{acl_natbib}

\clearpage
\appendix
% \section{Appendix}

\begin{table*}[!htbp]
\centering
\resizebox{\textwidth}{!}{%
\begin{tabular}{lp{45em}cp{7em}} \toprule
\# & \textbf{Utterances} & \textbf{Trust} & \textbf{Topic Discussing} \\
\midrule
1 & \underline{Therapist:} Hello, Peggy, how are you doing today? \newline
   \underline{Patient:} I'm okay. & 2.0 & Greeting \\
\midrule
2 & \underline{Therapist:} You're okay. Glad to hear that. Yesterday, when you were here at the agency, you took a few different tests. Remember that? \newline
   \underline{Patient:} I did. & 2.0 & Greeting \\
\midrule
3 & \underline{Therapist:} One of the tests you took is called the SCL 90, the symptom checklist 90. Do you remember taking that test? \newline
   \underline{Patient:} I think so. & 2.0 & Greeting \\
\midrule
4 & \underline{Therapist:} It's 90 items. And they are ranked. \newline
   \underline{Patient:} Asked me like true and false questions? & 2.0 & Greeting \\
\midrule
5 & \underline{Therapist:} Not true and false; they are numbers that you solve. \newline
   \underline{Patient:} Oh, okay. I do remember that. Okay. & 2.0 & Greeting \\
\midrule
6 & \underline{Therapist:} So I have the results from that test. I want to review some of those findings with you. \newline
   \underline{Patient:} Okay. All right. Sure. & 2.0 & Greeting \\
\midrule
7 & \underline{Therapist:} So this particular instrument has a number of what we refer to as subscales. For example, depression or anxiety. I'm going to review areas where you scored higher and discuss them with you. \newline
   \underline{Patient:} \colorbox{red!26}{Okay, are you recording this?} & 1.5 & Video Recording \\
\midrule
8 & \underline{Therapist:} No. \newline
   \underline{Patient:} \colorbox{red!26}{Are very worried about that?} & 1.5 & Video Recording \\
\midrule
9 & \underline{Therapist:} Is it what makes you think I'm recording? \newline
   \underline{Patient:} \colorbox{green!26}{Because that happens to me a lot.} & 2.0 & Video Recording \\
\midrule
10 & \underline{Therapist:} It happens to you a lot that you get recorded. \newline
    \underline{Patient:} \colorbox{green!26}{I think so. Yeah.} & 2.0 & Video Recording \\
\midrule
11 & \underline{Therapist:} Remember the last time it happened? \newline
    \underline{Patient:} Probably yesterday. & 2.0 & Video Recording \\
\midrule
12 & \underline{Therapist:} And you think you were recorded while you're taking this? \newline
    \underline{Patient:} Probably. Yeah. & 2.0 & Video Recording \\
\midrule
13 & \underline{Therapist:} That sounds worrisome. \colorbox{green!26}{That's a feeling you get a lot that you're being recorded.} \newline
    \underline{Patient:} \colorbox{green!26}{I worry about it a lot.} & 2.5 & Video Recording \\
\midrule
14 & \underline{Therapist:} You're certain that's happening? \newline
    \underline{Patient:} Yeah. & 2.5 & Video Recording \\
\midrule
15 & \underline{Therapist:} Where you're not being recorded now. \colorbox{green!26}{I know it probably feels like you are, but you're not.} \newline
    \underline{Patient:} Okay. \colorbox{green!26}{But I appreciate you bringing up that concern with me.} & 2.5 & Video Recording \\
\midrule
16 & \underline{Therapist:} In school, right? That would make sense. For example, if you had a percentile rank of 65\%, it means your score is equal to or higher than 65\% of people who took that test. Does that make sense? \newline
    \underline{Patient:} I think so. & 2.5 & Video Recording \\
\midrule
17 & \underline{Therapist:} Areas here on the SCL results where you scored higher than expected include the phobic anxiety scale. On that one, you were about 64th percentile. \newline
    \underline{Patient:} \colorbox{green!26}{So I have more phobic anxiety than most people.} & 2.5 & Anxiety \\
\midrule
18 & \underline{Therapist:} It does seem that way. \colorbox{green!26}{This construct measures fears like being afraid of going out or crowds.} Does that connect with you?

\underline{Patient:} \colorbox{green!26}{Yeah. Crowds make me nervous, and public speaking feels overwhelming.} & 3.0 & Anxiety \\
\midrule
19 & \underline{Therapist:} So it does appear you have phobic anxiety symptoms. That 64th percentile makes sense. \newline
    \underline{Patient:} I kind of get that. & 3.0 & Anxiety \\
\midrule
20 & \underline{Therapist:} On the paranoia scale, you were at the 84th percentile. \newline
    \underline{Patient:} \colorbox{red!26}{Is that why you were recording me?} & 3.0 & Paranoid \\
\bottomrule
\end{tabular}%
}
\caption{{\bf Dialogue Trust Table.} Highlights changes in trust during the interaction, focusing on key phrases in utterances that influence trust changes. Green indicates increased trust, while red shows a decrease.}
\label{tab:exampleDialog}
\end{table*}

\end{document}